%% file: main.tex
\documentclass[10pt,twocolumn,letterpaper]{article}

\usepackage[pagenumbers]{wacv} 


\definecolor{wacvblue}{rgb}{0.21,0.49,0.74}
\usepackage[pagebackref,breaklinks,colorlinks,allcolors=wacvblue]{hyperref}
\usepackage{multirow}
\usepackage{booktabs}
\usepackage{array}
\usepackage{graphicx}

\def\wacvPaperID{1822} 
\def\confName{WACV}
\def\confYear{2026}

\title{PEaRL: Pathway-Enhanced Representation Learning for Gene and Pathway Expression Prediction from Histology}

\author{
Sejuti Majumder$^{1}$, Saarthak Kapse$^{1}$, Moinak Bhattacharya$^{1}$,
Xuan Xu$^{2}$,\\ Alisa Yurovsky$^{1}$, Prateek Prasanna$^{1}$\\[2pt]
$^{1}$Department of Biomedical Informatics, Stony Brook University, NY, USA,\\
$^{2}$Department of Computer Science, Stony Brook University, NY, USA\\[2pt]
{\tt\small 
sejuti.majumder@stonybrook.edu, saarthak.kapse@stonybrook.edu, xuaxu@cs.stonybrook.edu,}\\
{\tt\small 
moinak.bhattacharya@stonybrook.edu,
alisa.yurovsky@stonybrookmedicine.edu,
}\\
{\tt\small 
prateek.prasanna@stonybrook.edu}
}

\begin{document}
\maketitle
\input{sec/0_abstract}

\input{sec/1_introduction}

\input{sec/2_relatedworks}
\input{sec/3_methods}

\input{sec/4_results}

\input{sec/5_conclusion}\\
\textbf{Acknowledgements.} This research was partially supported by National Institutes of Health (NIH) and National Cancer
Institute (NCI) grants 1R21CA258493-01A1, 1R01CA297843-01, 3R21CA258493-02S1, 1R03DE033489-01A1, and National Science Foundation (NSF) grant 2442053. The content is solely the responsibility of the authors and does not necessarily represent the official views of the National Institutes of Health.
{
    \small
    \bibliographystyle{ieeenat_fullname}
    \bibliography{main}
}

\input{sec/supplementary}

\end{document}

%% file: sec/0_abstract.tex
\begin{abstract}
Integrating histopathology with spatial transcriptomics (ST) provides a powerful opportunity to link tissue morphology with molecular function. Yet most existing multimodal approaches rely on a small set of highly variable genes, which limits predictive scope and overlooks the coordinated biological programs that shape tissue phenotypes. We present \textbf{PEaRL} (\textbf{P}athway \textbf{E}nhanced \textbf{R}epresentation \textbf{L}earning), a multimodal framework that represents transcriptomics through pathway activation scores computed with ssGSEA. By encoding biologically coherent pathway signals with a transformer and aligning them with histology features via contrastive learning, PEaRL reducesdimensionality, improves interpretability, and strengthens cross-modal correspondence.Across three cancer ST datasets—breast, skin, and lymph node—PEaRL consistently outperforms SOTA methods, yielding higher accuracy for both gene- and pathway-level expression prediction (up to 58.9\%  and 20.4\% increase in Pearson correlation coefficient compared to SOTA). These results demonstrate that grounding transcriptomic representation in pathways produces more biologically faithful and interpretable multimodal models, advancing computational pathology beyond gene-level embeddings.
\end{abstract}

%% file: sec/1_introduction.tex
\section{Introduction}
\label{sec:introduction}
Recent advancements in digital pathology have led to improvements in cell and tissue-level classification~\cite{cellvit,hovernet,tissue_classification,tissue_class2}, survival prediction~\cite{survival-2,survival-1}, and automated diagnostics~\cite{automated-diag}. Deep networks have demonstrated impressive capabilities in identifying phenotypic patterns associated with disease progression. However, despite these improvements, a fundamental challenge remains, clinical outcomes and disease states are typically determined by more than just morphological patterns. Tumor microenvironments are inherently complex, and direct correlations between tissue phenotype and molecular function remain elusive. This limitation motivates the need for approaches that integrate histopathology with molecular data to provide a more comprehensive view of disease mechanisms.\\
Transcriptomics~\cite{transcriptomics,transcriptomics-2} enables the genome-wide quantification of gene expression, providing insights into the molecular mechanisms underlying various biological processes. Advances in single-cell RNA sequencing~\cite{single-cell-1,single-cell2} have allowed for high-resolution profiling of gene expression at the individual cell level, improving the classification of distinct cell types and states. However, traditional bulk transcriptomics and single-cell sequencing approaches suffer from significant limitations. Bulk RNA sequencing averages gene expression across a heterogeneous population of cells, obscuring cell-specific expression patterns. Meanwhile, single-cell approaches, despite their high resolution, require tissue dissociation, leading to the loss of crucial spatial information necessary for understanding tissue organization and cell-cell interactions. \\
\begin{figure}
    \centering
    \includegraphics[width=1\linewidth]{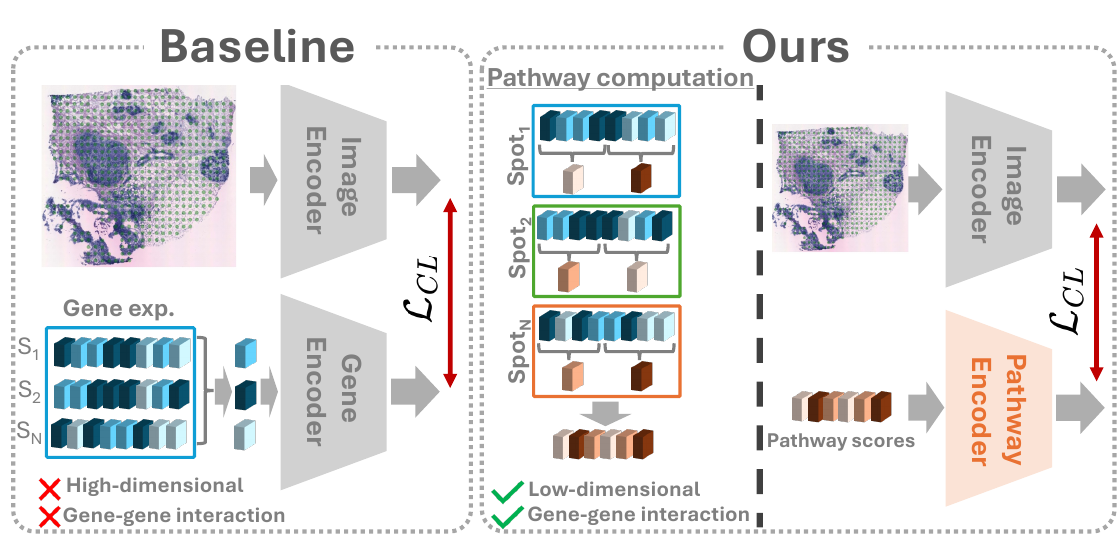}
    \caption{~\textbf{Multimodal Learning with PEaRL}. PEaRL considers gene-gene interactions via pathways which not only addresses the curse of high gene dimensionality but also acts as a molecular driver of tissue morphology leading to biologically meaningful representation learning between images and molecular data. We employ PEaRL for both gene and pathway expression prediction and also show its utility in survival analysis.}
    \label{fig:teaser_diagram}
\end{figure}
Spatial transcriptomics (ST) bridges the gap between transcriptomics
and histology by preserving the spatial context of gene expression within intact tissue sections~\cite{st-1,st-2}. Unlike bulk and single-cell RNA sequencing, ST enables the measurement of gene expression at precise tissue locations, facilitating the study of cell states and molecular interactions in their native microenvironments. This integration of spatial information with transcriptomic data makes ST ideal for multimodal learning with histopathology images. By aligning morphological patterns with spatially resolved gene expression, ST provides a unique opportunity to uncover the functional underpinnings of histopathological changes, ultimately improving disease diagnosis, prognosis, and therapeutic targeting.\\
Several computational methods have been developed to integrate histopathology images with transcriptomic data, yet this task remains fundamentally challenging. For the task of gene expression prediction from histology images, the overlapping but distinct information captured by these two modalities make the perfect reconstruction of transcriptomics profiles from morphology infeasible. Additionally, gene expression is both high-dimensional (often 15,000+ genes) and sparse, creating the curse of dimensionality that complicates model training. Early approaches such as STNet~\cite{STnet}, HisToGene~\cite{HisToGene}, His2ST~\cite{His2ST}, and THItoGene~\cite{thitogene} demonstrated the feasibility of mapping histology patches to gene expression using convolutional networks, transformers, or graph neural networks. However, these methods typically focused on predicting a small subset of highly variable or marker genes, which limited their predictive scope and often failed to capture the broader heterogeneity of transcriptomic signals. To overcome these constraints, more recent work has explored contrastive learning, which aligns histology and gene expression features in a shared latent space rather than directly regressing expression values. By encouraging embeddings from the same spatial location to cluster together, contrastive learning provides a more flexible way to capture cross-modal correspondences while reducing sensitivity to sparsity and noise in ST data. Methods such as HisToGene~\cite{HisToGene}, BLEEP~\cite{bleep}, and mclSTExp~\cite{mclstexp} highlight this trend, showing that contrastive objectives can improve alignment between image and transcriptomics and support more generalizable multimodal representations.\\
While gene-level encoding has been common practice in existing multimodal ST methods, it suffers from both technical and biological limitations. Using a small set of highly variable genes to mitigate high dimensionality limits both predictive scope and fails to capture the coordinated molecular programs that drive tissue morphology. Meanwhile, pathways, defined as sets of genes with known interactions and shared cellular functions, aggregate signals across all expressed genes into compact, biologically coherent modules. For example, genes such as the ~\textit{HLA-A},~\textit{HLA-B}, and ~\textit{STAT1}  may individually appear as highly variable and be used as markers of immune activity. However, considered in isolation, they do not reflect the broader immune processes at play. When examined together as part of the Hallmark Allograft Rejection pathway, these genes capture coordinated immune responses involving antigen presentation and interferon signaling, which more coherently explain the widespread lymphocyte infiltration observed in histology. Pathways provide a natural solution to these challenges. Compared to raw gene-level vectors, pathway scores reduce dimensionality while preserving the functional underpinnings of tissue organization and disease progression. Furthermore, because pathways represent interpretable biological processes, they constitute semantically meaningful tokens that can align more naturally with histological patterns, for instance, immune infiltration in histology corresponding to the activation of interferon or T-cell signaling pathways (see Fig.~\ref{fig:teaser_diagram}).\\
\textbf{Our approach.} We introduce \textbf{PEaRL} (\textbf{P}athway \textbf{E}nhanced \textbf{R}epresentation \textbf{L}earning), a multimodal framework that leverages pathway activation scores, computed via single-sample gene set enrichment analysis (ssGSEA), as the transcriptomic input. Instead of relying on individual genes, PEaRL encodes biologically coherent pathway signals using a transformer-based pathway encoder, while histology features are extracted with an image encoder. A contrastive objective aligns these two modalities in a shared latent space, enabling robust cross-modal correspondence. We evaluate PEaRL on three spatial transcriptomics oncology datasets, breast~\cite{her2st}, skin~\cite{ji2020multimodal}, and lymph node~\cite{meylan2022tertiary}, and demonstrate improved performance in both gene- and pathway-level expression prediction compared to existing methods. By grounding transcriptomic representation in pathways, PEaRL reduces dimensionality, enhances interpretability, and achieves more biologically faithful multimodal alignment.\\
Our contributions are threefold:
\begin{itemize}
\item We propose a pathway-based representation of spatial transcriptomics data, enabling biologically enriched embeddings for multimodal learning.
\item We design a multimodal framework that combines a transformer pathway encoder, an image encoder, and contrastive learning to jointly predict both pathway and gene expression from histology.
\item Through extensive experiments on multiple cancer datasets, we show that PEaRL consistently outperforms prior approaches in expression prediction and further demonstrates utility in downstream survival analysis.
\end{itemize}

%% file: sec/2_relatedworks.tex
\section{Related Work}

\subsection{Spatial Gene Expression Prediction}
Spatial transcriptomics (ST) has motivated a growing body of work on predicting gene expression directly from histology images. Early approaches such as STNet~\cite{STnet} leveraged CNN-derived features from H\&E patches to predict expression of a fixed gene panel, demonstrating the feasibility of image-to-transcriptome prediction. HisToGene~\cite{HisToGene} extended this idea with a Vision Transformer jointly encoding histology and spatial coordinates, enabling higher-resolution predictions. Hist2ST~\cite{His2ST} further decomposed the task into local, global, and neighborhood modules via ConvMixer, Transformer, and GraphSAGE. More recently, hybrid architectures such as THItoGene~\cite{thitogene} integrated multi-scale convolutional and attention-based blocks to better capture tumor microenvironmental cues.

Despite these advances, most regression-based models still rely on highly variable or marker genes, limiting predictive scope and failing to capture broader transcriptomic heterogeneity. To overcome these challenges, contrastive frameworks have been introduced. BLEEP~\cite{bleep} aligned histology and gene expression embeddings through a CLIP-style objective, imputing expression with k-nearest neighbors. MCLSTExp~\cite{mclstexp} unified histology, expression, and spatial coordinates into a single contrastive learning framework, modeling spots as “words” and tissue regions as “sentences.” These methods highlight the promise of contrastive objectives for robust multimodal alignment.

\subsection{Pathway-Based Representations}
Meanwhile, pathway-level representations have been increasingly explored in survival prediction as a way to overcome the high dimensionality and noise of gene-level features. By aggregating genes into functional modules, pathway scores provide compact and interpretable signals that link molecular programs to prognosis. For example, SurvPath~\cite{jaume2024modeling} models dense interactions between pathways and histology patches using a multimodal transformer, improving patient risk stratification. Cox-Path~\cite{ma2024cox} integrates multi-omics data into a pathway-informed graph convolutional network, capturing pathway–pathway interactions before Cox regression. More recently, PIBD~\cite{zhang2024prototypical} addresses redundancy in multimodal survival prediction by disentangling common and modality-specific features across pathways and images. Beyond general frameworks, disease-specific studies such as PPAR-related prognostic modeling in hepatocellular carcinoma (HCC)~\cite{kimura2012ppar} further demonstrate how pathway activity can define subgroups with distinct outcomes and therapeutic responses. Together, these works highlight pathways as natural semantic units for survival analysis and motivate their use in multimodal frameworks like PEaRL.

\subsection{Histopathology Representation Learning}
Parallel to ST-specific efforts, the field of histopathology has witnessed rapid progress in image representation learning. Weakly supervised whole-slide image (WSI) analysis has been the dominant paradigm, with multiple instance learning (MIL) frameworks widely applied to tasks such as cancer classification~\cite{sudharshan2019multiple,xu2012multiple} and survival prediction~\cite{survival-1,survival-2}. With the rise of self-supervised learning (SSL), approaches such as SimCLR, MoCo, and DINO have been adapted to WSIs, giving rise to pathology foundation models like CTransPath~\cite{wang2022transformer}, HIPT~\cite{chen2022scaling}, and UNI~\cite{uni}, which generalize across diverse downstream tasks.

In the multimodal domain, CLIP~\cite{radford2021clip} inspired vision–language pretraining has been adapted to pathology through PLIP~\cite{huang2023visual}, CONCH~\cite{conch}, and Prov-GigaPath~\cite{Prov-Giga}, which align histology images with captions or molecular data. These models have established new benchmarks in retrieval, classification, and subtyping, highlighting the utility of large-scale multimodal pretraining. Importantly, the availability of ST data has opened opportunities for aligning histology with molecular profiles at spatial resolution, motivating specialized frameworks such as BLEEP and mclSTExp that explicitly target expression prediction.

%% file: sec/3_methods.tex
\section{Methods}
\label{sec:methods}

\begin{figure*}
    \centering
    \includegraphics[width=1\linewidth]{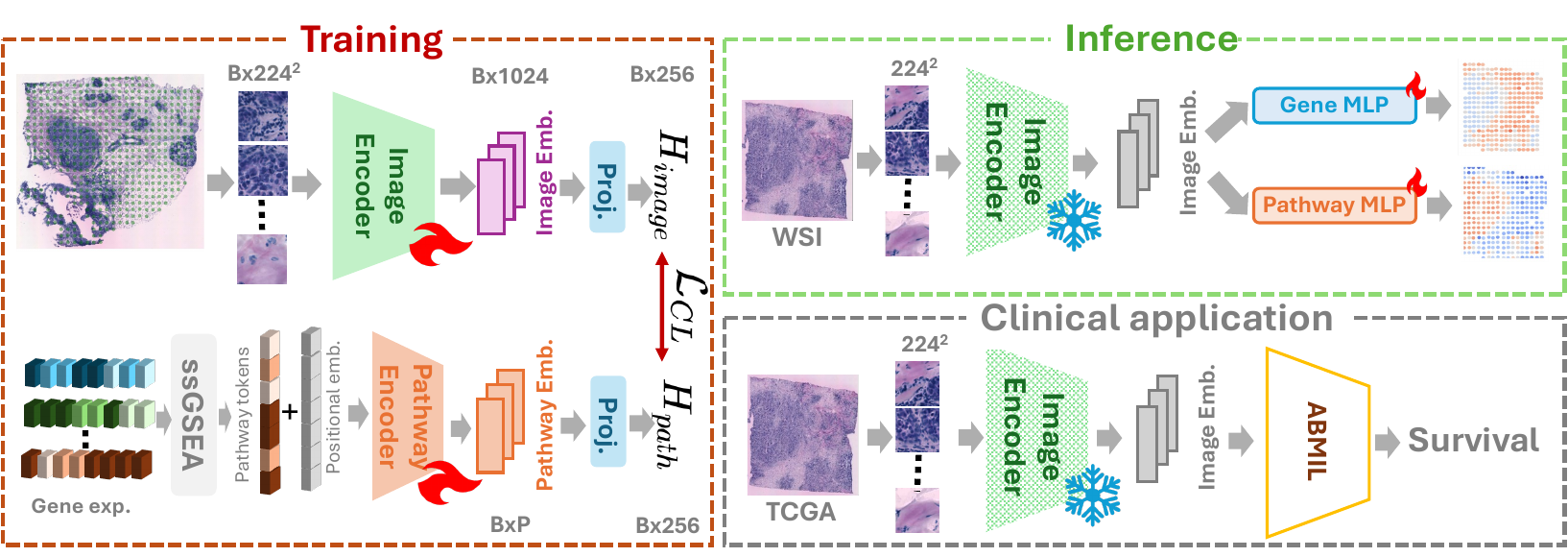}
    
    \caption{Overview of PEaRL framework. The framework begins with extracting patches corresponding to spatial transcriptomics spots and feeding them to the image encoder to get image embeddings. We compute the pathway scores from the high dimensional gene expression data corresponding to the spatial transcriptomics spots and feed it as an input to the transformer based pathway encoder along with the positional embeddings. The image embeddings and pathway embeddings are aligned via contrastive learning.The learned image embeddings are then used for downstream gene and pathway expression prediction.}
    \label{fig:framework}
\end{figure*}

\noindent\textbf{Overview.} PEaRL employs multimodal learning to integrate histological image features with ST data for gene and pathway expression prediction (Fig.~\ref{fig:framework}). The training framework consists of two key components: the \textit{pathway encoder} and the \textit{vision encoder}, which generate embeddings for pathways and histological images, respectively. These embeddings are aligned using contrastive learning to establish meaningful cross-modal representations.



\subsection{Pathway Encoder}
The pathway encoder integrates biological context by leveraging pathway scores derived from gene expression. A biological pathway can be defined as a network of molecular interactions~\cite{kuenzi2020census} where genes ($G$) function together in a coordinated manner to drive cellular processes such as metabolism, signaling, and regulation that, in turn, influence the cellular structure and tissue organization. 

To compute pathway scores, we apply single-sample Gene Set Enrichment Analysis (ssGSEA)~\cite{ssgsea} directly on the expression matrix. Unlike GSEA~\cite{gsea}, which compares gene expression between two predefined groups (e.g., diseased vs. healthy samples) to determine pathway enrichment at a population level, ssGSEA computes an enrichment score for each sample independently by ranking genes and summing their weighted expression values. This enables pathway activity assessment at a spot level, capturing spatial heterogeneity across tissue sections. Given a pathway \( P_i \) containing \( G_{P_i} \) genes, where \(P \) denotes the number of pathways considered, the Normalized Enrichment Score (NES) for a spot \( s \) is computed as:
\begin{equation}
NES_s(P_i) = \frac{ES_s(P_i)}{\mathbb{E}\left[\,|ES_s(P_{\text{null}})|\,\right]}
\end{equation}

where $ES_s(P_i)$ is the enrichment score of pathway $P_i$ in spot $s$, 
computed from the ranked gene expression profile of that spot; 
$P_{\text{null}}$ are random gene sets of the same size as $P_i$; 
and the denominator normalizes by the expected score from these null sets. 
Thus, $NES_s(P_i)$ serves as the pathway score for pathway $P_i$ in spot $s$.

Thus, for each point \( s \), we compute a pathway score vector \( x_{\text{s}} \). The dimensionality of this vector corresponds to the number of pathways included for a given dataset:
\begin{equation}
x_{\text{s}} \in \mathbb{R}^{P}
\end{equation}

\paragraph{Transformer as the pathway encoder.}
For a batch of \(N\) spots, let \(X \in \mathbb{R}^{N \times P}\) denote the ssGSEA pathway score matrix. We encode spatial position using globally normalized coordinates such that it maps the coordinates of the spots across the dataset to a common reference point. This global normalization ensures that a batch of $N$ spots can contain positions from different slides while remaining comparable in a shared coordinate space.  For raw spot coordinates \(C \in \mathbb{R}^{N \times 2}\) (x,y), we compute:
\begin{equation}
\tilde{C} = \frac{C - \mu}{\sigma}
\end{equation}
with mean \(\mu \in \mathbb{R}^{2}\) and standard deviation \(\sigma \in \mathbb{R}^{2}\) estimated over all spots in the split (train or validation). A learnable positional encoder \(\phi: \mathbb{R}^{2} \rightarrow \mathbb{R}^{P}\) (implemented as a MLP) maps \(\tilde{C}\) to \(\phi(\tilde{C}) \in \mathbb{R}^{N \times P}\). The transformer input is the sum:
\begin{equation}
H_0 = X + \phi(\tilde{C}) \;\in\; \mathbb{R}^{N \times P}
\end{equation}

We use a two-layer Transformer encoder with standard multi-head self-attention (MHSA). For each head \(h\),
\begin{equation}
Q_h = H_0 W^{(h)}_Q,\quad K_h = H_0 W^{(h)}_K,\quad V_h = H_0 W^{(h)}_V
\end{equation}
and attention is computed as
\begin{equation}
\mathrm{Attn}_h(H_0) = \mathrm{softmax}\!\left(\frac{Q_h K_h^{\top}}{\sqrt{d_k}}\right) V_h
\end{equation}
where \(W^{(h)}_Q, W^{(h)}_K, W^{(h)}_V \in \mathbb{R}^{P \times d_k}\) and \(d_k\) is the per-head key/query dimension. Outputs from all heads are concatenated and projected, followed by a position-wise feed-forward network with residual connections and layer normalization. The encoder output:
\begin{equation}
Z \in \mathbb{R}^{N \times P}
\end{equation}
provides pathway embeddings that integrate pathway activity with spatial context.

After obtaining the output from transformer, we apply a projection head to map each spot's pathway embedding into a 256-dimensional space for multimodal alignment and contrastive learning:
\begin{equation}
H_{path} = MLP(Z) \in \mathbb{R}^{N\times {256}}
\end{equation}

\subsection{Vision Encoder}
To extract histological features, we employ UNI v1~\cite{uni}, a pre-trained foundational model trained on 100k whole slide images, 
making it an ideal choice for encoding rich morphological features from H\&E stained tissue sections. To process histopathology images, we first extract \( 224 \times 224 \)
patches from whole slide images, ensuring that each patch corresponds to a ST spot. Each patch \(I_i\) in the batch of $N$ spots/patches is passed through the UNI encoder to obtain a 1024 dimensional image embedding. Since our spot embeddings are 256 dimensional, we apply projection head to align the image feature dimension with the spot embedding space. Additionally we fine-tune the last 4 layers of UNI, allowing the model to adapt to this domain.

\begin{align}
Z_{\text{image}} &= \text{UNI}(I) \in \mathbb{R}^{N \times 1024} \\
H_{\text{image}} &= \text{MLP}(Z_{\text{image}}) \in \mathbb{R}^{N \times 256}
\end{align}

\subsection{Training and Inference}

\paragraph{Stage 1: Contrastive pretraining.}
We learn a shared 256-dimensional space by aligning image and pathway embeddings with a symmetric contrastive loss. The
\(H_{\text{image}} \in \mathbb{R}^{N \times 256}\) and
\(H_{\text{path}} \in \mathbb{R}^{N \times 256}\)
are then L2-normalized for a batch of \(N\) matched image–pathway pairs, followed by computing similarity logits as follows:
\begin{equation}
\label{eq:contrastive-sim}
\mathbf{S} \;=\; \frac{H_{\text{image}}\, H_{\text{path}}^{\top}}{\tau} \;\in\; \mathbb{R}^{N \times N},
\end{equation}
where \(\tau>0\) is a learnable temperature. With identity targets (treating diagonal elements as positives for contrastive learning), we optimize row-wise and column-wise cross-entropy (CE):
\begin{equation}
\label{eq:row-ce}
\mathcal{L}_{\text{img}\rightarrow\text{path}} \;=\; \frac{1}{N}\sum_{i=1}^{N}
\mathrm{CE}\!\left(\mathrm{softmax}(\mathbf{S}_{i:}),\, i\right),
\end{equation}
\begin{equation}
\label{eq:col-ce}
\mathcal{L}_{\text{path}\rightarrow\text{img}} \;=\; \frac{1}{N}\sum_{j=1}^{N}
\mathrm{CE}\!\left(\mathrm{softmax}(\mathbf{S}_{:j}),\, j\right).
\end{equation}
The symmetric contrastive objective is
\begin{equation}
\label{eq:cl-total}
\mathcal{L}_{\text{CL}} \;=\; \tfrac{1}{2}\big(\mathcal{L}_{\text{img}\rightarrow\text{path}}
+ \mathcal{L}_{\text{path}\rightarrow\text{img}}\big).
\end{equation}

\paragraph{Stage 2: Supervised prediction heads (frozen backbone).}
After contrastive pretraining, we freeze the backbone and train two lightweight MLP heads (\(f_{\text{path}}\) and \(f_{\text{gene}}\)) on top of the image embeddings. For a batch, let
\(H_{\text{image}} \in \mathbb{R}^{N \times 256}\) denote these embeddings.
We predict pathway scores and gene expression with two heads as follows:
\begin{equation}
\label{eq:head-path}
\hat{y}_{\text{path}} \;=\; f_{\text{path}}(H_{\text{image}}) \;\in\; \mathbb{R}^{N \times P},
\end{equation}
\begin{equation}
\label{eq:head-gene}
\hat{y}_{\text{gene}} \;=\; f_{\text{gene}}(H_{\text{image}}) \;\in\; \mathbb{R}^{N \times g},
\end{equation}

where $g$ denotes the number of highly variable genes (HVGs). We train these heads with mean-squared error (MSE) against ground truth pathway scores
\(y_{\text{path}} \in \mathbb{R}^{N \times P}\) and gene expression
\(y_{\text{gene}} \in \mathbb{R}^{N \times g}\):
\begin{equation}
\label{eq:mse-path}
\mathcal{L}_{\text{path}} \;=\; \frac{1}{N \cdot P}\,\lVert \hat{y}_{\text{path}} - y_{\text{path}} \rVert_2^2,
\end{equation}
\begin{equation}
\label{eq:mse-gene}
\mathcal{L}_{\text{gene}} \;=\; \frac{1}{N \cdot g}\,\lVert \hat{y}_{\text{gene}} - y_{\text{gene}} \rVert_2^2.
\end{equation}
The supervised objective for head training is the sum:
\begin{equation}
\label{eq:heads-total}
\mathcal{L}_{\text{sup}} \;=\; \,\mathcal{L}_{\text{path}} \;+\; \,\mathcal{L}_{\text{gene}},
\end{equation}

\paragraph{Inference.}
At test time, an input patch \(I_{\text{test}}\) is passed through the trained image encoder and projection head to obtain \(h_{\text{image}}^{\text{test}} \in \mathbb{R}^{256}\). The pathway and gene predictions are then produced directly by the trained MLP heads:
\begin{equation}
\label{eq:infer-path}
\hat{y}_{\text{path}}^{\text{test}} \;=\; f_{\text{path}}\!\left(h_{\text{image}}^{\text{test}}\right),
\end{equation}
\begin{equation}
\label{eq:infer-gene}
\hat{y}_{\text{gene}}^{\text{test}} \;=\; f_{\text{gene}}\!\left(h_{\text{image}}^{\text{test}}\right).
\end{equation}

%% file: sec/4_results.tex
\section{Experiments and Results}
We evaluate PEaRL on gene expression and pathway expression, along with survival analysis as a downstream task.

\begin{table*}[!ht]
\centering
\caption{Performance comparison on \textbf{Gene Expression} prediction across datasets. Results are mean $\pm$ std. Best per column in \textbf{bold}.}
\label{tab:gene_prediction}
\resizebox{\textwidth}{!}{
\begin{tabular}{cccccccccc}
\toprule
\multirow{2}{*}{\textbf{Method}}  & \multicolumn{3}{c}{\textbf{Breast}} & \multicolumn{3}{c}{\textbf{Skin}} & \multicolumn{3}{c}{\textbf{Lymph}} \\
\cmidrule(lr){2-4} \cmidrule(lr){5-7} \cmidrule(lr){8-10}
 & \textbf{PCC}  $\uparrow$ & \textbf{MSE}  $\downarrow$ & \textbf{MAE}  $\downarrow$ & \textbf{PCC}  $\uparrow$ & \textbf{MSE}  $\downarrow$ & \textbf{MAE}  $\downarrow$ & \textbf{PCC}  $\uparrow$ & \textbf{MSE}  $\downarrow$ & \textbf{MAE}  $\downarrow$ \\
\midrule
STNet     & 0.0010 $\pm$ 0.0008 & 0.5682 $\pm$ 0.0270 & 0.5186 $\pm$ 0.0152 & 0.0002 $\pm$ 0.0010 & 0.7161 $\pm$ 0.0508 & 0.5755 $\pm$ 0.0204 & 0.0016 $\pm$ 0.0018 & 0.3035 $\pm$ 0.0099 & 0.3846 $\pm$ 0.0038 \\
His2ST    & 0.0916 $\pm$ 0.0550 & 0.1899 $\pm$ 0.0115 & 0.2997 $\pm$ 0.0088 & 0.1197 $\pm$ 0.0574 & 0.2041 $\pm$ 0.0076 & 0.3334 $\pm$ 0.0038 & 0.1480 $\pm$ 0.0441 & 0.1428 $\pm$ 0.0039 & 0.2317 $\pm$ 0.0038 \\
BLEEP     & 0.1057 $\pm$ 0.0521 & 0.2009 $\pm$ 0.0199 & 0.2956 $\pm$ 0.0200 & 0.0696 $\pm$ 0.0284 & 0.2241 $\pm$ 0.0131 & 0.3354 $\pm$ 0.0096 & 0.0622 $\pm$ 0.0249 & 0.1814 $\pm$ 0.0239 & 0.2473 $\pm$ 0.0149 \\
MCLSTExp  & 0.4654 $\pm$ 0.0493 & 0.1096 $\pm$ 0.0069 & 0.2162 $\pm$ 0.0065 & 0.2810 $\pm$ 0.0227 & 0.1595 $\pm$ 0.0072 & 0.2888 $\pm$ 0.0055 & 0.1248 $\pm$ 0.0245 & 0.1842 $\pm$ 0.0095 & 0.2515 $\pm$ 0.0083 \\
\midrule
PEaRL (ours) & \textbf{0.5868 $\pm$ 0.0359} & \textbf{0.0732 $\pm$ 0.0033} & \textbf{0.1828 $\pm$ 0.0043} & \textbf{0.3756 $\pm$ 0.0148} & \textbf{0.1405 $\pm$ 0.0071} & \textbf{0.2726 $\pm$ 0.0073} & \textbf{0.2352 $\pm$ 0.0145} & \textbf{0.1425 $\pm$ 0.0058} & \textbf{0.2278 $\pm$ 0.0031} \\
\bottomrule
\end{tabular}
}
\end{table*}

\begin{table*}[!ht]
\centering
\caption{Performance comparison on \textbf{Pathway Expression} prediction across datasets. Results are mean $\pm$ std. Best per column in \textbf{bold}.}
\label{tab:pathway_prediction}
\resizebox{\textwidth}{!}{
\begin{tabular}{cccccccccc}
\toprule
\multirow{2}{*}{\textbf{Method}} & \multicolumn{3}{c}{\textbf{Breast}} & \multicolumn{3}{c}{\textbf{Skin}} & \multicolumn{3}{c}{\textbf{Lymph}} \\
\cmidrule(lr){2-4} \cmidrule(lr){5-7} \cmidrule(lr){8-10}
 & \textbf{PCC}  $\uparrow$ & \textbf{MSE}  $\downarrow$& \textbf{MAE}  $\downarrow$& \textbf{PCC}  $\uparrow$ & \textbf{MSE}  $\downarrow$ & \textbf{MAE}  $\downarrow$ & \textbf{PCC}  $\uparrow$ & \textbf{MSE}  $\downarrow$ & \textbf{MAE}  $\downarrow$ \\
\midrule
STNet     & 0.0006 $\pm$ 0.0031 & 0.1469 $\pm$ 0.0037 & 0.3052 $\pm$ 0.0040 & 0.0008 $\pm$ 0.0023 & 0.1972 $\pm$ 0.0040 & 0.3538 $\pm$ 0.0035 & 0.0018 $\pm$ 0.0022 & 0.1197 $\pm$ 0.0191 & 0.2596 $\pm$ 0.0197 \\
His2ST    & 0.2008 $\pm$ 0.0270 & 0.0045 $\pm$ 0.0003 & 0.0506 $\pm$ 0.0014 & 0.1071 $\pm$ 0.0501 & 0.0046 $\pm$ 0.0002 & 0.0520 $\pm$ 0.0016 & 0.0324 $\pm$ 0.0280 & 0.0040 $\pm$ 0.0003 & 0.0480 $\pm$ 0.0014 \\
BLEEP     & 0.0560 $\pm$ 0.0714 & 0.0036 $\pm$ 0.0004 & 0.0449 $\pm$ 0.0033 & 0.0958 $\pm$ 0.0703 & 0.0054 $\pm$ 0.0009 & 0.0540 $\pm$ 0.0042 & 0.0610 $\pm$ 0.0317 & 0.0062 $\pm$ 0.0009 & 0.0602 $\pm$ 0.0052 \\
MCLSTExp  & 0.4197 $\pm$ 0.0465 & 0.0020 $\pm$ 0.0001 & 0.0333 $\pm$ 0.0010 & 0.3306 $\pm$ 0.0386 & 0.0034 $\pm$ 0.0003 & 0.0430 $\pm$ 0.0015 & \textbf{0.2396 $\pm$ 0.0155} & 0.0029 $\pm$ 0.0002 & 0.0405 $\pm$ 0.0016 \\
\midrule
PEaRL (ours) & \textbf{0.5055 $\pm$ 0.0271} & \textbf{0.0017 $\pm$ 0.0001} & \textbf{0.0314 $\pm$ 0.0010} & \textbf{0.3523 $\pm$ 0.0336} & \textbf{0.0029 $\pm$ 0.0002} & \textbf{0.0404 $\pm$ 0.0012} & 0.2247 $\pm$ 0.0353 & \textbf{0.0028 $\pm$ 0.0003} & \textbf{0.0401 $\pm$ 0.0018} \\
\bottomrule
\end{tabular}
}
\end{table*}

\subsection{Dataset and Implementation}

\begin{figure*}[h]
    \centering
    \includegraphics[width=1\linewidth]{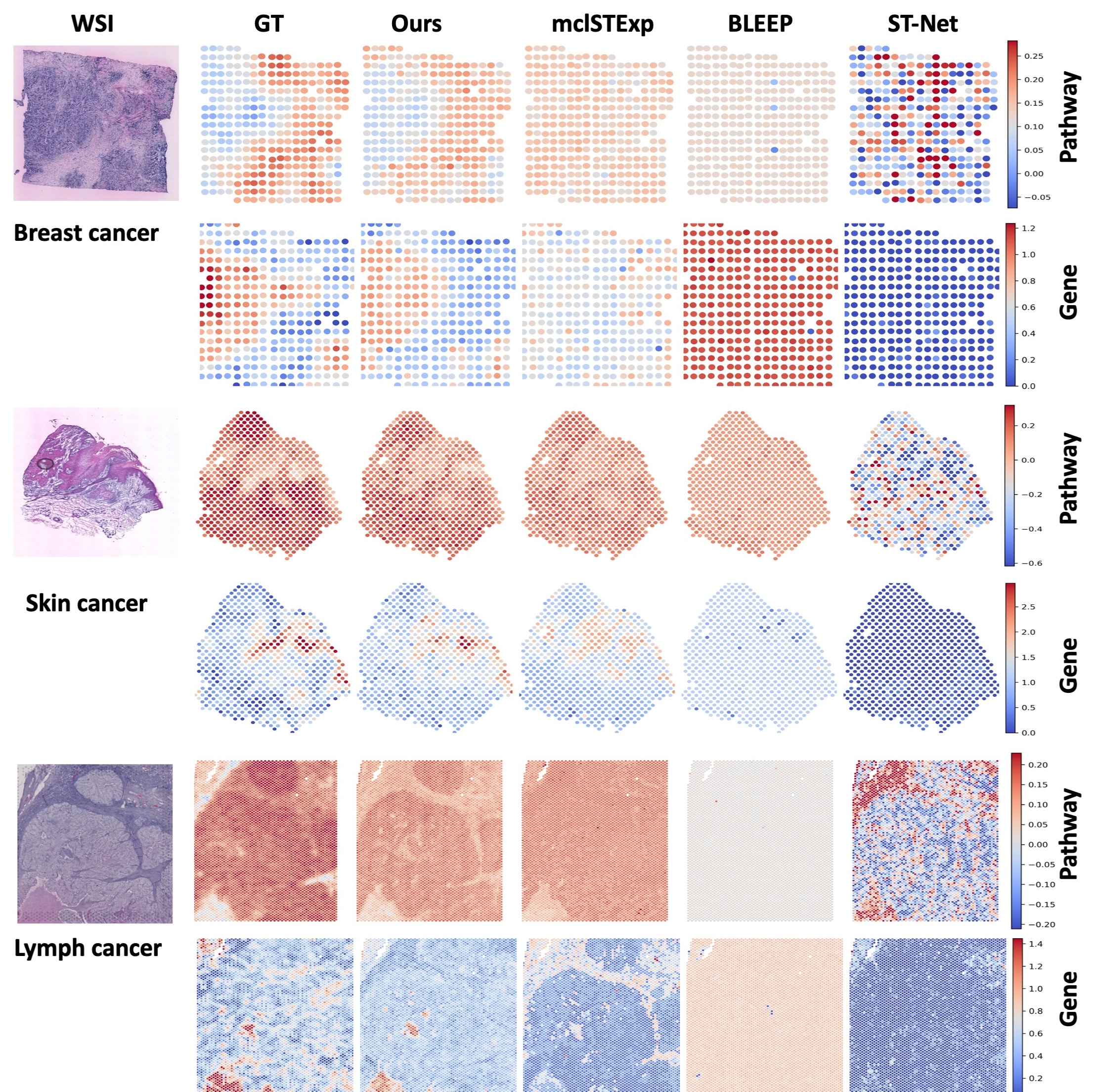}
    \caption{Visualization of expression predictions of different pathways and their corresponding genes across three datasets compared with three baseline methods. We show Hallmark\_allograft\_rejection pathway and its gene HLA-DMB for the breast cancer case; Hallmark\_epithelial\_mesenchymal transition pathway and its gene QSOX1 for the skin cancer case; Reactome\_ABC\_family\_proteins\_mediated\_transport and its gene DERL3 for the lymph carcinoma case.}
    \label{fig:heatmaps}
\end{figure*}

\textbf{Datasets.} 
We evaluate PEaRL using three cancer-focused spatial transcriptomics datasets curated from the HEST-1k Library~\cite{jaume2025hest}, a standardized multimodal resource of paired histology and gene expression profiles. The breast cancer dataset~\cite{her2st} comprises 36 tissue sections with 13,620 spatial spots. The skin cancer dataset~\cite{ji2020multimodal} contains 12 samples with 8,671 spots, while the lymph cancer dataset~\cite{meylan2022tertiary} consists of 24 samples with 74,220 spots.\\
\textbf{Data preprocessing and pathway annotation.} 
 We first filtered out genes detected in fewer than 1,000 spots, followed by total-count normalization to 10,000 counts per spot and logarithmic transformation. We applied 8-neighbor smoothing to reduce spot-level noise like high sparsity while preserving local spatial continuity in tissue structure. Highly variable genes (HVGs) were then selected using Scanpy~\cite{wolf2018scanpy}, retaining the top 1,000 HVGs ($g$) for each dataset. For pathway annotation, we integrated gene sets from Reactome~\cite{reactome} and MSigDB~\cite{msigdb}, and quantified pathway activity at the single-spot level using single-sample gene set enrichment analysis (ssGSEA)~\cite{ssgsea}. This yielded $P=775$ pathways for the breast cancer dataset, $P=1,100$ pathways for the lymph dataset, and $P=609$ pathways for the skin dataset.  \\
\textbf{Baselines.} 
We benchmark PEaRL against several state-of-the-art models for histology–transcriptomics integration. \textbf{ST-Net}~\cite{he2020integrating} employs a CNN encoder trained with regression loss to predict spot-level gene expression. \textbf{Hist2ST}~\cite{zeng2022spatial} combines a ConvMixer encoder with a GraphSAGE module and optimizes with regression loss. \textbf{BLEEP}~\cite{bleep} adopts a ResNet/ViT encoder trained with a contrastive loss to align image and expression embeddings. \textbf{mclSTExp}~\cite{mclstexp} employs CNN encoders with spatial integer encoding, also trained with a contrastive loss.\\
\textbf{Evaluation metrics.} 
Model performance on gene and pathway expression prediction is evaluated using Pearson correlation coefficient (PCC), mean squared error (MSE), and mean absolute error (MAE) between predicted and ground-truth values as per previous studies~\cite{he2020integrating,zeng2022spatial,bleep} For survival analysis, we report the concordance index (C-index), which measures the agreement between predicted risk scores and observed survival outcomes~\cite{chen2021multimodal,wang2009rna,xu2023multimodal}.\\
\textbf{Implementations.} 
All models were trained using 5-fold cross-validation. PEaRL was optimized using AdamW with a learning rate of \(1 \times 10^{-4}\), weight decay of \(1 \times 10^{-3}\), and trained for up to 100 epochs with early stopping (patience 15). The pathway encoder is a Transformer with two attention layers ($h=8$ heads, $d_k = 64$ dimensions each). Experiments were conducted on one NVIDIA Quadro RTX 8000 GPU.

\subsection{Quantitative Results}

\textbf{Gene expression prediction.}
As shown in Table~\ref{tab:gene_prediction}, across all three datasets, PEaRL consistently improves over baseline methods in terms of PCC. On the breast dataset, PEaRL achieves a PCC of 0.5868 compared to 0.4654 for the strongest baseline (MCLSTExp), corresponding to a 26.1\% relative gain. On the skin dataset, PEaRL obtains 0.3756 vs.\ 0.2810 for MCLSTExp, a 33.7\% improvement. The largest margin is observed on the lymph dataset, where PEaRL reaches 0.2352 compared to 0.1480 for His2ST, yielding a 58.9\% relative increase.\\
\textbf{Pathway expression prediction.}
For pathway-level targets, PEaRL achieves notable gains on the breast dataset, improving from PCC of 0.4197 (MCLSTExp) to 0.5055, a 20.4\% relative increase (see Table~\ref{tab:pathway_prediction}). On skin, PEaRL improves from PCC of 0.3306 to 0.3523 (6.6\%), while on lymph the performance is slightly lower. Overall, the strongest benefit is observed in the breast dataset.\\
\textbf{Survival analysis.} 
We further evaluated prognostic utility on TCGA-BRCA whole-slide images using an ABMIL framework~\cite{ilse2018attention} trained with a Cox partial likelihood objective. All models were trained and evaluated under a 5-fold cross-validation setting. As shown in Table~\ref{tab:survival}, PEaRL achieved the highest concordance index (C-index = 0.659 $\pm$ 0.027), surpassing both purely histology-based encoders such as ResNet50 (0.513 $\pm$ 0.031) and UNI (0.588 $\pm$ 0.046), as well as multimodal baselines such as mclstexp (0.612 $\pm$ 0.029), which incorporate only gene-level information. These results demonstrate that multimodal pathway-vision models capture prognostic trends more effectively than embeddings derived from histology alone or from gene-level representations. 
\begin{table}[ht]
\centering
\caption{Survival analysis on \textbf{TCGA-BRCA} (C-index). Results are mean $\pm$ std.}
\label{tab:survival}
\begin{tabular}{lcc}
\toprule
\textbf{Pretraining Modality} & \textbf{Image Encoders} & \textbf{c-index} \\
\midrule
ImageNet-1K  & ResNet50   & 0.513 $\pm$ 0.031 \\

WSI & UNI        & 0.588 $\pm$ 0.046 \\
WSI+Gene & mclstexp   & 0.612 $\pm$ 0.029 \\
WSI+Pathway & PEaRL      & \textbf{0.659 $\pm$ 0.027} \\
\bottomrule
\end{tabular}
\end{table}

\subsection{Ablations}
We performed ablation experiments on the breast cancer dataset to disentangle the contributions of different model components and design choices, summarized in Table~\ref{tab:ablation_study}.\\
\begin{figure*}[ht]
    \centering
    \includegraphics[width=1\linewidth]{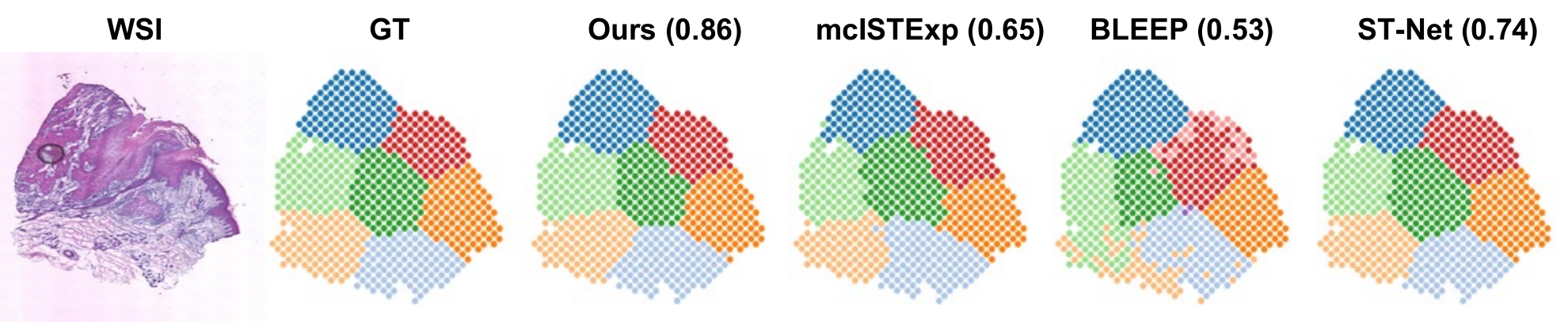}
    \caption{Visualization of the Leiden clusterings for ground truth (GT) and predicted gene expressions (for a skin cancer sample) using PEaRL and other baselines. ARI index shown in (.).}
    \label{fig:leiden_cluster}
\end{figure*}
\begin{table*}[ht]
\centering
\caption{Ablation analysis on \textbf{Breast} dataset for Gene and Pathway prediction. Results are mean $\pm$ std.}
\label{tab:ablation_study}
\resizebox{\textwidth}{!}{
\begin{tabular}{ccccccc}
\toprule
\multirow{2}{*}{\textbf{Breast}} & \multicolumn{3}{c}{\textbf{Gene Prediction}} & \multicolumn{3}{c}{\textbf{Pathway Prediction}} \\
\cmidrule(lr){2-4} \cmidrule(lr){5-7}
 & \textbf{PCC} $\uparrow$ & \textbf{MSE} $\downarrow$ & \textbf{MAE} $\downarrow$ & \textbf{PCC} $\uparrow$ & \textbf{MSE} $\downarrow$ & \textbf{MAE} $\downarrow$ \\
\midrule
Image encoder (UNI) only   & 0.1882 $\pm$ 0.014 & 0.4316 $\pm$ 0.025 & 0.4760 $\pm$ 0.015 & 0.1814 $\pm$ 0.010 & 0.0971 $\pm$ 0.003 & 0.2513 $\pm$ 0.006 \\
Pathway encoder only & 0.2154 $\pm$ 0.016 & 0.3757 $\pm$ 0.025 & 0.4125 $\pm$ 0.018 & 0.4265 $\pm$ 0.026 & 0.0118 $\pm$ 0.000 & 0.0880 $\pm$ 0.001 \\

\midrule

PEaRL (UNI default)        & \textbf{0.5868 $\pm$ 0.0359} & \textbf{0.0732 $\pm$ 0.0033} & \textbf{0.1828 $\pm$ 0.0043}  & \textbf{0.5055 $\pm$ 0.0271} & \textbf{0.0017 $\pm$ 0.0001} & \textbf{0.0314 $\pm$ 0.0010} 
\\

\midrule

PEaRL w/ ResNet50   & 0.3729 $\pm$ 0.0511 & 0.1351 $\pm$ 0.0074 & 0.2430 $\pm$ 0.0074 & 0.3337 $\pm$ 0.0500 & 0.0023 $\pm$ 0.0002 & 0.0360 $\pm$ 0.0014 
\\
PEaRL w/ DenseNet121 & 0.4833 $\pm$ 0.0634 & 0.1090 $\pm$ 0.0024 & 0.2180 $\pm$ 0.0019 & 0.3822 $\pm$ 0.0549 & 0.0022 $\pm$ 0.0002 & 0.0349 $\pm$ 0.0014 
\\
PEaRL w/ ViT-B        & 0.3573 $\pm$ 0.0389 & 0.1380 $\pm$ 0.0056 & 0.2459 $\pm$ 0.0062  & 0.3303 $\pm$ 0.0344 & 0.0023 $\pm$ 0.0001 & 0.0356 $\pm$ 0.0006
\\

\bottomrule
\end{tabular}
}
\end{table*}
\noindent\textbf{Single-modality encoders.}  
We here evaluated unimodal models. Using the image encoder alone allows us to test how much predictive signal can be derived purely from histology, independent of molecular priors. Conversely, using only the pathway encoder to predict gene expression directly evaluates the strength of pathway-derived features. While the pathway-only setup achieves strong performance on pathway prediction by construction, it is also more effective than the image encoder for gene expression prediction (PCC = 0.2154 vs.\ 0.1882). This indicates that while histology carries complementary information, pathway-level features are inherently more predictive. This highlights the critical role of incorporating pathway priors beyond morphology alone. \\
\textbf{Different image encoders.}  
Next, we examined the impact of varying the architecture of the image encoder (Table~\ref{tab:ablation_study}). We compared three types of models: (i) conventional CNNs such as ResNet50 and DenseNet121, (ii) a Vision Transformer (ViT), and (iii) UNI, a foundation model pretrained on large-scale histopathology images. Note that apart from UNI, all three encoders were trained from scratch with our PEaRL framework. Across both \emph{gene expression} and \emph{pathway prediction} tasks, UNI consistently achieved the best performance, showcasing the advantage of pretraining on domain-specific histopathology corpora. CNNs and ViTs, while competitive, lagged behind UNI, demonstrating that generic backbones are less effective.

\subsection{Qualitative Results}
\textbf{Breast cancer pathway and gene visualization.}  
Fig.~\ref{fig:heatmaps} shows visualizations of the Hallmark allograft rejection pathway~\cite{le2002multiple,marino2016allorecognition} and the corresponding HLA-DMB gene in the breast cancer dataset. This pathway captures immune-related activity, which is highly relevant in breast tumor microenvironments. The HLA-DMB gene is an important component of the antigen presentation machinery and plays a role in immune surveillance. Our pathway maps reveal spatially distinct immune activity within the tissue, which is reflected consistently in the gene-level predictions. Together, the pathway- and gene-level visualizations demonstrate that PEaRL can capture not only global immune signaling but also its spot-level variation, producing coherent and biologically meaningful maps compared to SOTA. Additional heatmaps can be seen in supplementary Fig. 9-10.

\noindent\textbf{Leiden clustering analysis.}  
We use Leiden clustering, a graph-based community detection algorithm, to partition spots into coherent groups~\cite{anuar2021comparison,traag2019louvain}. By clustering the predicted expression maps, we assess whether the model reconstructs biologically meaningful tissue structures. To evaluate cluster agreement, we use the Adjusted Rand Index (ARI), which measures similarity between predicted and ground-truth clusters while correcting for random chance. 
 PEaRL achieves higher ARI scores than baselines (Fig.~\ref{fig:leiden_cluster}). Leiden clustering of the breast and lymph cancer samples are shown in supplementary Figs. 7-8. We also show correlation plots for three datasets in the supplementary Figs. 5-6.

%% file: sec/5_conclusion.tex
\section{Conclusion}

In this work, we introduced PEaRL, a novel multimodal framework that integrates histology images with pathway-informed transcriptomic representations to predict gene and pathway expression from histology images. Our experiments across multiple cancer ST datasets demonstrate that PEaRL consistently outperforms existing baselines in both gene and pathway prediction tasks, and in the capture of prognostic information. Furthermore, qualitative analyses revealed that PEaRL recovers biologically meaningful spatial patterns, maintains co-expression structure, and produces clusters that align with histological regions. Together, these results highlight PEaRL’s ability to bridge the gap between molecular and morphological views of cancer. Future work will extend this framework to pan-cancer datasets, incorporate additional modalities, and explore clinical applications such as biomarker discovery and treatment response prediction.

%% file: sec/supplementary.tex
\clearpage

~\section{Supplementary Material}
In the supplementary material we begin with the correlation plots (Figures 5 and 6), and show the leiden clusters for the breast and lymph cancer samples (Figures 7 and 8). Finally, we also show additional pathway and gene heatmaps for the breast,skin,and lymph cancer samples (Figures 9 and 10).

\begin{figure*}
    \centering
    \includegraphics[width=1\linewidth]{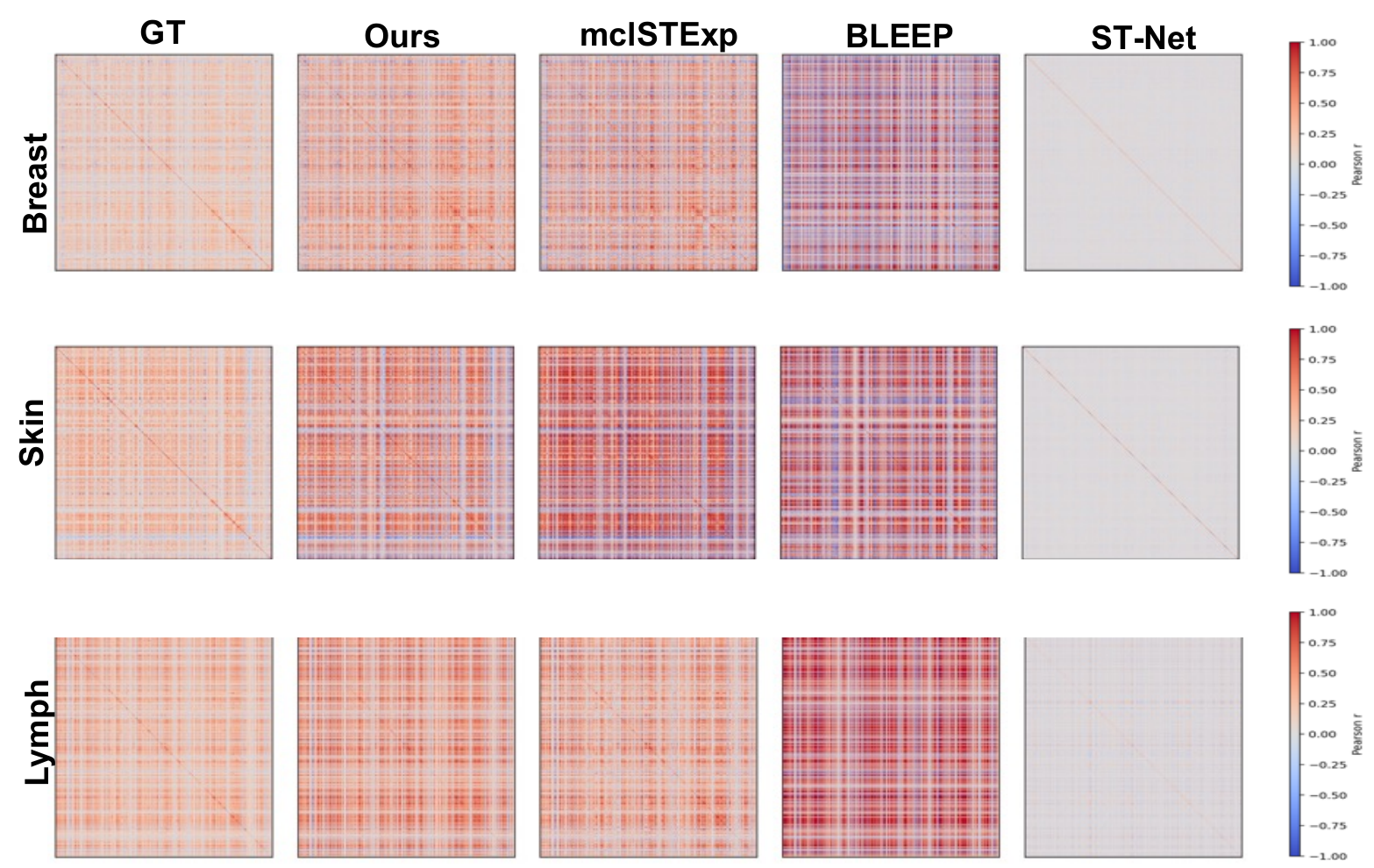}
    \caption{Visualization of the pathway-pathway correlation plots across the three datasets and comparison of PEaRL with baseline models}
    \label{fig:placeholder}
\end{figure*}

\begin{figure*}
    \centering
    \includegraphics[width=1\linewidth]{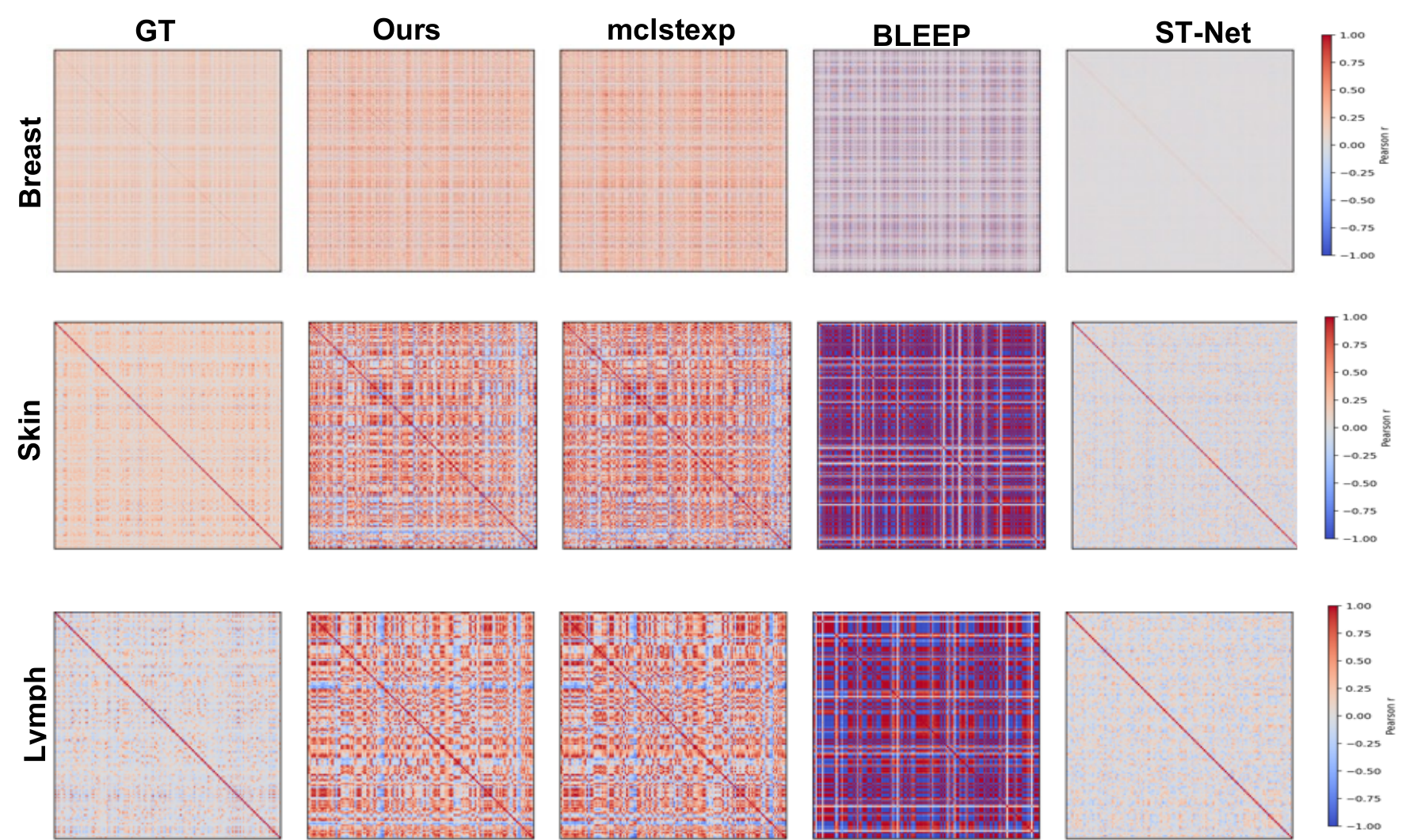}
    \caption{Visualization of the gene-gene correlation plots across the three datasets and comparison of PEaRL with baseline models}
    \label{fig:placeholder}
\end{figure*}

\begin{figure*}
    \centering
    \includegraphics[width=1\linewidth]{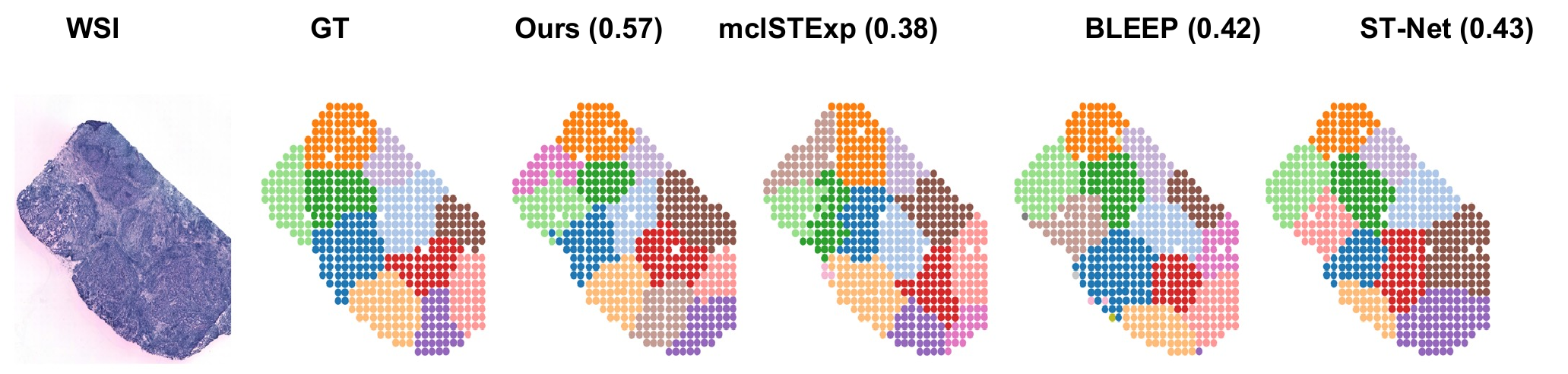}
    \caption{Visualization of leiden clusterings for ground truth and predicted gene expressions for the breast cancer dataset. ARI index shown in (.)}
    \label{fig:placeholder}
\end{figure*}

\begin{figure*}
    \centering
    \includegraphics[width=1\linewidth]{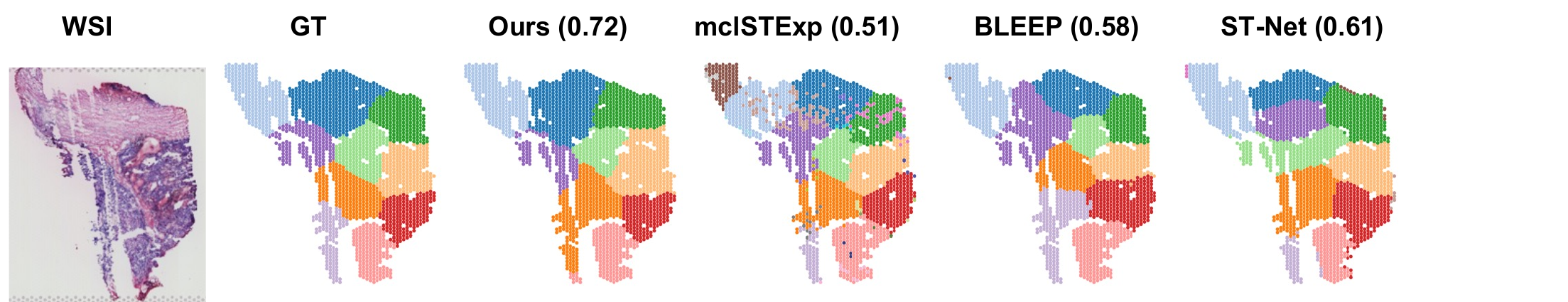}
    \caption{Visualization of leiden clusterings for ground truth and predicted gene expressions for the lymph cancer sample. ARI index shown in (.)}
    \label{fig:placeholder}
\end{figure*}

\begin{figure*}
    \centering
    \includegraphics[width=1\linewidth]{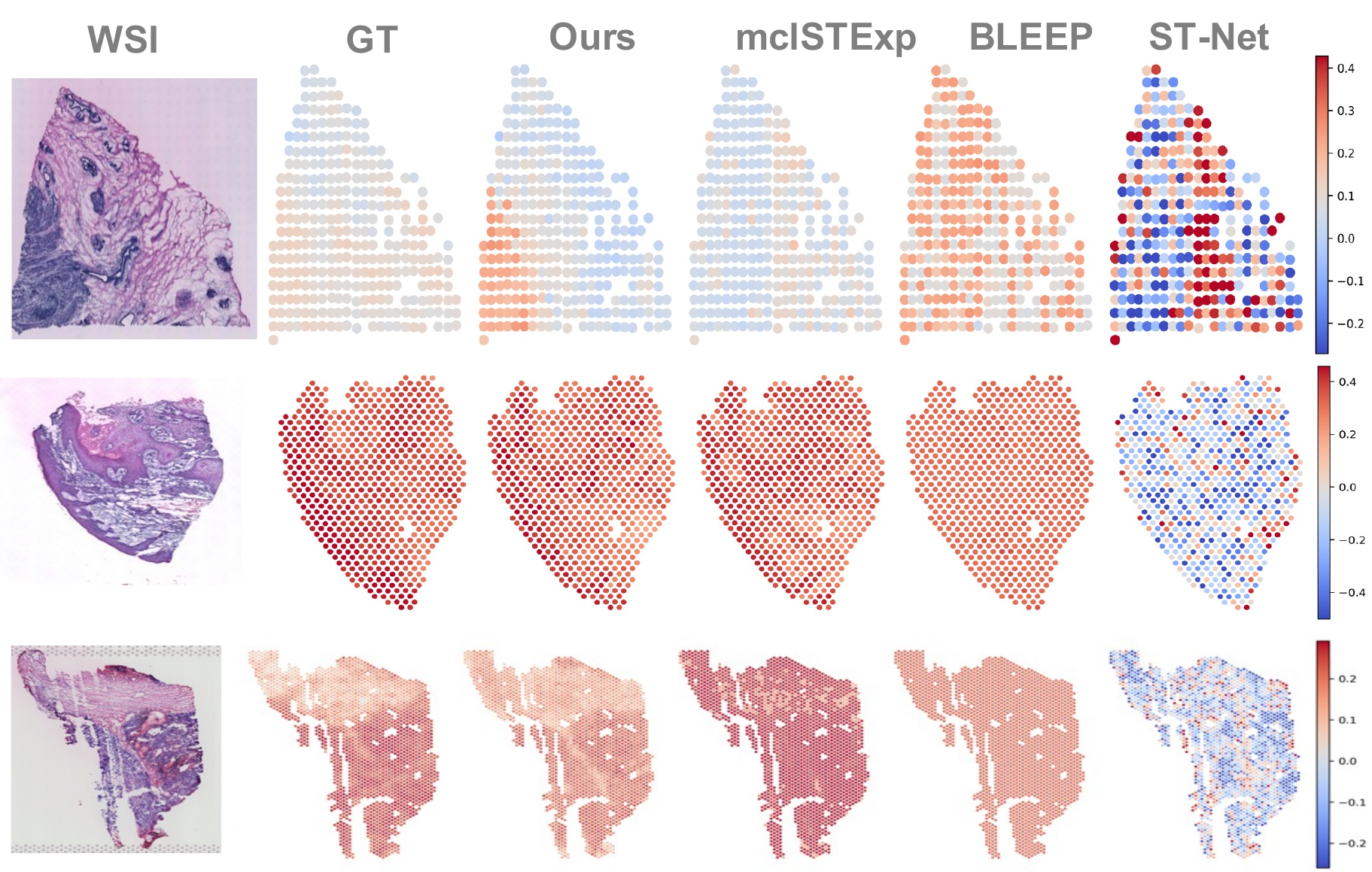}
    \caption{Visualization of pathway heatmaps across the three datasets.We show the \texttt{Hallmark\_Myc\_targets\_v1} pathway for the breast sample, 
    \texttt{Reactome\_Eukaryotic\_Translation\_Initiation} for the skin sample, 
    and \texttt{Reactome\_ABC\_family\_of\_proteins\_mediated\_transport} for the lymph cancer sample.}
    \label{fig:placeholder}
\end{figure*}

\begin{figure*}
    \centering
    \includegraphics[width=1\linewidth]{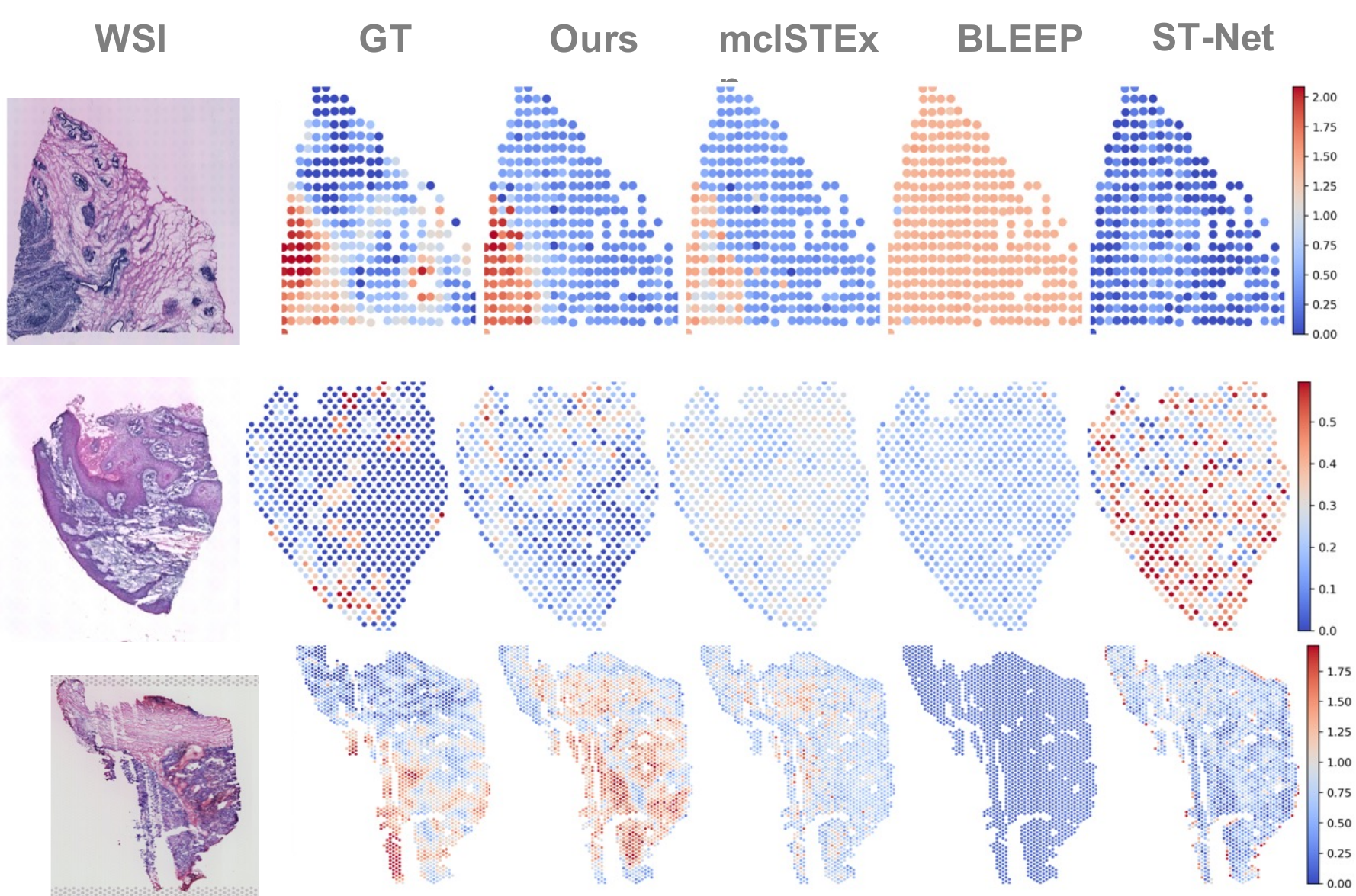}
    \caption{Visualization of the gene heatmaps across the three datasets.We show genes SSBP1, E1F4EBP1, and PKP2 for the breast, skin, and the lymph cancer samples respectively.}
    \label{fig:placeholder}
\end{figure*}